\documentclass[10pt,a4paper,onecolumn]{IEEEtran}
\usepackage[lmargin=2cm, rmargin=2cm, tmargin=2cm, bmargin=2cm]{geometry}
\usepackage[latin1]{inputenc}
\usepackage[T1]{fontenc}
\usepackage{ccaption}
\usepackage[T1]{fontenc}
\usepackage{graphicx}
\usepackage{fancyhdr}
\usepackage{helvet}
\usepackage{floatflt}
\usepackage{colortbl}
\usepackage{fancyvrb}
\usepackage{wrapfig}
\usepackage[english]{babel}
\usepackage[round,authoryear]{natbib}
\usepackage[english]{babel}
\usepackage{graphicx}
\usepackage{subfigure}
\usepackage{amsfonts}
\usepackage{amsmath}
\usepackage{theorem}
\usepackage{ae}
\usepackage{bm}

\newtheorem{theorem}{Theorem}
\newtheorem{assumption}{Assumption}

\newcommand{\tz}{\tilde{z}}
\newcommand{\dtz}{\dot{\tilde{z}}}

\title{A neural network based controller for underwater robotic vehicles}
\author{Josiane Maria Macedo Fernandes, Marcelo Costa Tanaka,\\
Raimundo Carlos Silvério Freire Júnior, Wallace Moreira Bessa}
\date{}

\begin{document}

\maketitle

\abstract{
\noindent
Due to the enormous technological improvements obtained in the last decades 
it is possible to use robotic vehicles for underwater exploration. This work describes the development of 
a dynamic positioning system for remotely operated underwater vehicles based. The adopted approach is 
developed using Lyapunov Stability Theory and enhanced by a neural network based algorithm for uncertainty 
and disturbance compensation. The performance of the proposed control scheme is evaluated by means of 
numerical simulations.}

\section{INTRODUCTION}

The control system is one of the most important peaces of a Remotely Operated underwater Vehicle (ROV), 
and its characteristics (advantages and disadvantages) play an essential role when one have to choose a 
vehicle for a specific mission. These vehicles have been substituting the divers in the accomplishment 
of tasks that offer risks to the human life. In this way, ROVs have been used thoroughly in the research 
of sub phenomena and in assembly, inspection, and repair of offshore structures. During the execution 
of a certain task with the robotic vehicle, the operator needs to monitor and control a series of parameters. 
If some of these parameters, as for instance the position and attitude of the vehicle, could be attended 
automatically by a control system, the teleoperation of the ROV can be enormously facilitated.

Unfortunately, the problem of designing accurate positioning systems for underwater robotic vehicles still 
challenges many engineers and researchers interested in this particular branch of engineering science. A 
growing number of papers dedicated to the position and orientation control of such vehicles confirms the 
necessity of the development of a controller, that could deal with the inherent nonlinear system dynamics, 
imprecise hydrodynamic coefficients, and external disturbances. It has already been shown \citep{yuh2,goheen1} 
that, in the case of underwater vehicles, the traditional control methodologies are not the most suitable 
choice and cannot guarantee the required tracking performance. 

Intelligent control, on the other hand, has proven to be a very attractive approach to cope with uncertain nonlinear systems 
\citep{tese,rsba2010,cobem2005,nd2012,Bessa2014,Bessa2017,Bessa2018,Bessa2019,Deodato2019,Lima2018,Lima2020,Lima2021,Tanaka2013}. 
By combining nonlinear control techniques, such as feedback linearization or sliding modes, with adaptive intelligent algorithms, 
for example fuzzy logic or artificial neural networks, the resulting intelligent control strategies can deal with the nonlinear 
characteristics as well as with modeling imprecisions and external disturbances that can arise.

In this work, a nonlinear controller scheme with a neural network based compensation scheme is employed for 
the dynamic positioning of underwater vehicles. Based on a Lyapunov stability theory , the convergence 
properties of the closed-loop system is analytically proven. Numerical results are also provided to 
confirm the control system efficacy.

\section{NONLINEAR CONTROLLER DESIGN}

Consider a class of $n^\mathrm{th}$-order nonlinear systems:

\begin{equation}
x^{(n)}=f(\mathbf{x})+b(\mathbf{x})u+d
\label{eq:system}
\end{equation}

\noindent
where $u$ is the control input, the scalar variable $x$ is the output of interest, $x^{(n)}$ is the 
$n$-th time derivative of $x$, $\mathbf{x}=[x,\dot{x},\ldots,x^{(n-1)}]$ is the system state vector, 
$d$ represents external disturbances and unmodeled dynamics, and $f,b:\mathbb{R}^n\to\mathbb{R}$ are 
both nonlinear functions.

As demonstrated by \citet{rsba2010}, adaptive fuzzy algorithms can be properly combined with nonlinear 
controllers in order to improve the trajectory tracking of uncertain nonlinear systems. It has also been 
shown that such strategies are suitable for a variety of applications ranging from remotely operated 
underwater vehicles \citep{raas2008,raas2010} to chaos control \citep{csf2009}.

The proposed control problem is to ensure that the state vector $\mathbf{x}$ will follow a desired 
trajectory $\mathbf{x}_d=[x_d,\dot{x}_d,\ldots,x^{(n-1)}_d]$ in the state space. Regarding the 
development of the control law, the following assumptions should also be made:

\begin{assumption}
The state vector $\mathbf{x}$ is available.
\label{hp:stat}
\end{assumption}
\begin{assumption}
The desired trajectory $\mathbf{x}_d$ is once differentiable in time. Furthermore, every element of 
vector $\mathbf{x}_d$, as well as $x^{(n)}_d$, is available and with known bounds.
\label{hp:traj}
\end{assumption}

Let $\tilde{x}=x-x_d$ be defined as the tracking error in the variable $x$, and 

\begin{displaymath}
\mathbf{\tilde{x}}=\mathbf{x}-\mathbf{x}_d=[\tilde{x},\dot{\tilde{x}},\ldots,\tilde{x}^{(n-1)}]
\end{displaymath}

\noindent 
as the tracking error vector. Now, consider a combined tracking error measure: 

\begin{equation}
\varepsilon=\mathbf{c^\mathrm{T}\tilde{x}}
\label{eq:measu}
\end{equation}

\noindent
where $\mathbf{c}=[c_{n-1}\lambda^{n-1},\ldots,c_1\lambda,c_0]$, $\lambda$ is positive constant and 
$c_i$ states for binomial coefficients, i.e.,

\begin{equation}
c_i=\binom{n-1}{i}=\frac{(n-1)!}{(n-i-1)!\:i!}\:,\quad\quad i=0,1,\ldots,n-1 
\label{eq:binom}
\end{equation}

\noindent
which makes $c_{n-1}\lambda^{n-1}+\cdots+c_1\lambda+c_0$ a Hurwitz polynomial. 

From Eq.~(\ref{eq:binom}), it can be easily verified that $c_0=1$, for $\forall n\ge1$. Thus, for 
notational convenience, the time derivative of $\varepsilon$ will be written in the following form:

\begin{equation}
\dot{\varepsilon}=\mathbf{c^\mathrm{T}\dot{\tilde{x}}}
=\tilde{x}^{(n)}+\mathbf{\bar{c}^\mathrm{T}\tilde{x}}
\label{eq:dmeasu}
\end{equation}

\noindent
where $\mathbf{\bar{c}}=[0,c_{n-1}\lambda^{n-1},\ldots,c_1\lambda]$.

Based on Assumptions~\ref{hp:stat}~and~\ref{hp:traj}, the following control law can be proposed:

\begin{equation}
u=\frac{1}{b}(-f-d+x^{(n)}_d-\mathbf{\bar{c}^\mathrm{T}\tilde{x}}-\kappa\varepsilon)
\label{eq:law}
\end{equation}

\noindent
where $\kappa$ is a strictly positive constant. 

The boundedness and convergence properties of the closed-loop system are established in the following
theorem.

\begin{theorem}
\label{th:stab}
Consider the nonlinear system (\ref{eq:system}) and Assumptions~\ref{hp:stat}--\ref{hp:traj}. Then, the controller 
defined by (\ref{eq:law}) ensures the exponential convergence of the tracking error, i.e., $\mathbf{\tilde{x}\to0}$ 
as $t\to\infty$.
\end{theorem}

\noindent
{\bf Proof:} Let a positive definite Lyapunov function candidate $V$ be defined as

\begin{equation}
V(t)=\frac{1}{2}\varepsilon^2
\label{eq:liap}
\end{equation}

\noindent
Thus, the time derivative of $V$ is

\begin{displaymath}
\dot{V}(t)=\varepsilon\dot{\varepsilon}=(\tilde{x}^{(n)}+\mathbf{\bar{c}^\mathrm{T}\tilde{x}})\varepsilon
=(x^{(n)}-x^{(n)}_d+\mathbf{\bar{c}^\mathrm{T}\tilde{x}})\varepsilon
=\big[f+b\,u+d-x^{(n)}_d+\mathbf{\bar{c}^\mathrm{T}\tilde{x}}\big]\varepsilon
\end{displaymath}

\noindent
By applying the proposed control law (\ref{eq:law}), one has

\begin{equation}
\dot{V}(t)=-\kappa\varepsilon^2
\label{eq:liap_p}
\end{equation}

\noindent
which implies $\varepsilon\to0$ as $t\to\infty$. 

\noindent
From the definition of limit, it means that for every $\xi>0$ there is a corresponding number $\tau$ such that 
$|\varepsilon|<\xi$ whenever $t>\tau$. According to Eq.~(\ref{eq:measu}) and considering that $|\varepsilon|<
\xi$ may be rewritten as $-\xi<\varepsilon<\xi$, one has 

\begin{equation}
-\xi<c_0\tilde{x}^{(n-1)}+c_1\lambda\tilde{x}^{(n-2)}+\cdots+c_{n-2}\lambda^{n-2}\dot{\tilde{x}}
+c_{n-1}\lambda^{n-1}\tilde{x}<\xi
\label{eq:ebounds}
\end{equation}

\noindent
Multiplying (\ref{eq:ebounds}) by $e^{\lambda t}$ and noting that

\begin{displaymath}
\displaystyle
\frac{d^{n-1}}{dt^{n-1}}(\tilde{x}e^{\lambda t})=\left(c_0\tilde{x}^{(n-1)}+c_1\lambda\tilde{x}^{(n-2)}
+\cdots+c_{n-2}\lambda^{n-2}\dot{\tilde{x}}+c_{n-1}\lambda^{n-1}\tilde{x}\right)e^{\lambda t}
\end{displaymath}

\noindent
one has 

\begin{equation}
-\xi e^{\lambda t}<\frac{d^{n-1}}{dt^{n-1}}(\tilde{x}e^{\lambda t})<\xi e^{\lambda t}\\ 
\label{eq:mbounds}
\end{equation}

\noindent
Thus, integrating (\ref{eq:mbounds}) $n-1$ times between $0$ and $t$ gives

\begin{multline}
-\frac{\xi}{\lambda^{n-1}}e^{\lambda t}+\left(\left|\frac{d^{n-2}}{dt^{n-2}}(\tilde{x}e^{\lambda t})
\right|_{t=0}+\frac{\xi}{\lambda}\right)\frac{t^{n-2}}{(n-2)!}+\cdots+\left(|\tilde{x}(0)|+
\frac{\xi}{\lambda^{n-1}}\right)\le\tilde{x}e^{\lambda t}\le\frac{\xi}{\lambda^{n-1}}e^{\lambda 
t}+\\+\left(\left|\frac{d^{n-2}}{dt^{n-2}}(\tilde{x}e^{\lambda t})\right|_{t=0}-\frac{\xi}{\lambda}
\right)\frac{t^{n-2}}{(n-2)!}+\cdots+\left(|\tilde{x}(0)|-\frac{\xi}{\lambda^{n-1}}\right)
\label{eq:int_n-1}
\end{multline}

\noindent
Furthermore, dividing (\ref{eq:int_n-1}) by $e^{\lambda t}$, it can be easily verified that the values of 
$\tilde{x}$ can be made arbitrarily close to $0$ (within a distance $\xi$) by taking $t$ sufficiently large 
(larger than $\tau$), i.e., $\tilde{x}\to0$ as $t\to\infty$. Now, considering the $(n-2)^\mathrm{th}$ integral
of (\ref{eq:mbounds}), dividing again by $e^{\lambda t}$ and considering that $\tilde{x}$ converges to zero, 
it follows that $\dot{\tilde{x}}\to0$ as $t\to\infty$. The same procedure can be successively repeated 
until the convergence of each component of the tracking error vector is achieved: $\mathbf{\tilde{x}\to0}$ 
as $t\to\infty$.\hfill$\square$ \vspace*{12pt}

\section{VEHICLE DYNAMICS MODEL}

A reasonable model to describe the underwater vehicle's dynamical behavior must include the 
rigid-body dynamics of the vehicle's body and a representation of the surrounding fluid dynamics. 
Such a model must be composed of a system of ordinary differential equations, to represent 
rigid-body dynamics, and partial differential equations to represent both tether and fluid 
dynamics. 

In order to overcome the computational problem of solving a system with this degree of complexity, in the 
majority of publications \citep{raas2010,raas2008,antonelli1,hoang1,smallwood2,hsu1,kiriazov1,yoerger2} 
a lumped-parameters approach is employed to approximate vehicle's dynamical behavior. 

\begin{figure}[htb]
\centering
\includegraphics[width=0.7\textwidth]{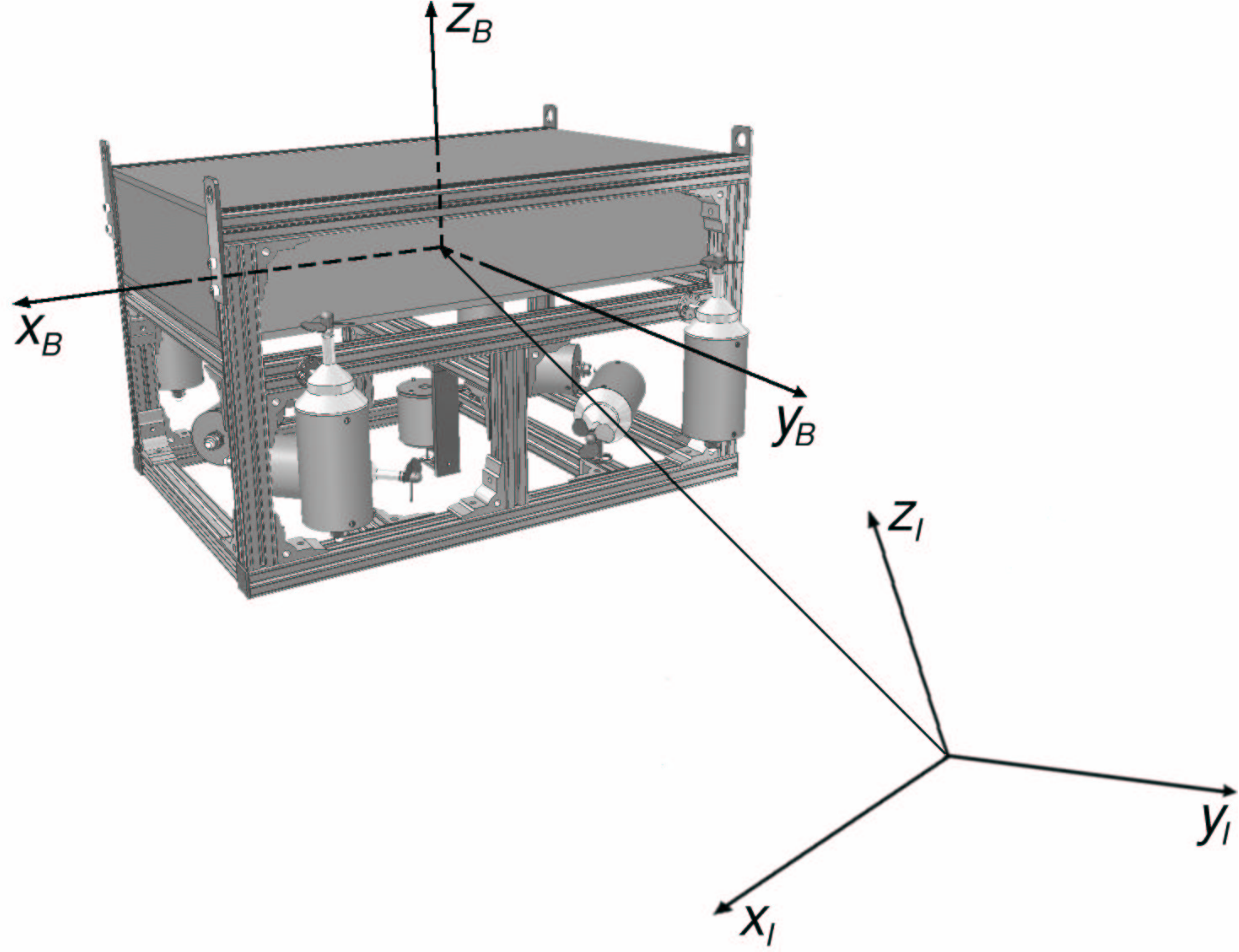}
\caption{Underwater vehicle with both inertial and body-fixed reference frames.} 
\label{fg:fig1}
\end{figure}

The equations of motion for underwater vehicles can be presented with respect to an inertial 
reference frame or with respect to a body-fixed reference frame, Fig.~\ref{fg:fig1}. On this 
basis, the equations of motion for underwater vehicles can be expressed, with respect to the 
body-fixed reference frame, in the following vectorial form:\\

\begin{equation}
\mathbf{M}\bm{\dot{\nu}}+\mathbf{k}(\bm{\nu})+\mathbf{h}(\bm{\nu})+\mathbf{g}(\mathbf{x})
+\mathbf{d}=\bm{\tau} 
\label{eq:movb}
\end{equation}

\noindent
where $\bm{\nu}=[\upsilon_x,\upsilon_y,\upsilon_z,\omega_x,\omega_y,\omega_z]$ is the vector of 
linear and angular velocities in the body-fixed reference frame, $\mathbf{x}=[x,\:y,\:z,\alpha,
\beta,\gamma]$ represents the position and orientation with respect to the inertial reference 
frame, $\mathbf{M}$ is the inertia matrix, which accounts not only for the rigid-body inertia 
but also for the so-called hydrodynamic added inertia, $\mathbf{k}(\bm{\nu})$ is the vector of 
generalized Coriolis and centrifugal forces, $\mathbf{h}(\bm{\nu})$ represents the hydrodynamic 
quadratic damping, $\mathbf{g}(\mathbf{x})$ is the vector of generalized restoring forces (gravity 
and buoyancy), $\mathbf{d}$ stands for occasional disturbances, and $\bm{\tau}$ is the vector of 
control forces and moments. 

It should be noted that in the case of remotely operated underwater vehicles (ROVs), the metacentric 
height is sufficiently large to provide the self-stabilization of roll ($\alpha$) and pitch ($\beta$) 
angles. This particular constructive aspect also allows the order of the dynamic model to be reduced 
to four degrees of freedom, $\mathbf{x}=[x,\:y,\:z,\gamma]$, and the vertical motion (heave) to be 
decoupled from the motion in the horizontal plane. This simplification can be found in the majority 
of works presented in the specialized literature \citep{hoang1,zanoli1,guo1,hsu1,kiriazov1,pinto1,%
cunha2,yoerger2}. Thus, the positioning system of a ROV can be divided in two different parts: Depth 
control (concerning variable $z$), and control in the horizontal plane (variables $x$, $y$ and 
$\gamma$).

Another important issue in the case of ROVs is the disturbance force caused by the umbilical (or 
tether cable). The umbilical can be treated as a continuum, discretized with the finite element 
method or modeled as multibody system \citep{bevilacqua1,pinto1}. However, the adoption of any of 
these approaches requires a computational effort that would be prohibitive for on-line estimation 
of the control action. A common way to surmount this limitation is to consider the forces and 
moments exerted by the tether as random, and incorporate them into the vector $\mathbf{d}$. 

On this basis, considering that the restoring forces can be passively compensated \citep{kiriazov1}, 
the most relevant hydrodynamic forces and moments acting on ROVs are discussed in the following 
subsections.

\subsection{Hydrodynamic forces}

Remotely operated underwater vehicles typically operate with velocities never exceeding 2 m/s.
Consequently, the hydrodynamic forces ($F_h$) can be approximated using the {\it Morison equation} 
\citep{newman1}:

\begin{equation}
F_h = \frac{1}{2}C_D A \rho\,v|v| + C_M\rho\nabla\dot{v} + \rho\nabla\dot{v}_w
\label{eq:morison}
\end{equation}

\noindent
where $v$ and $\dot{v}$ are, respectively, the relative velocity and the relative acceleration between 
rigid-body and fluid, $\dot{v}_w$ is the acceleration of underwater currents, $A$ is a reference area, 
$\rho$ is the fluid density, $\nabla$ is the fluid's displaced volume, $C_D$ and $C_M$ are coefficients 
that must be obtained experimentally.

The last term of Eq.~(\ref{eq:morison}) is the so-called {\it Froude-Kryloff force} and will not be
considered in this work due the fact, that at normal working depths, the acceleration of the underwater
currents is negligible. In this way, the coefficient $C_M\rho\nabla$ of the second term will be called
hydrodynamic added mass. The first term represents the nonlinear hydrodynamic quadratic damping.
Experimental tests \citep{kleczka1} show that Morison equation describes with sufficient accuracy the 
hydrodynamic effects due to the relative motion between rigid-bodies and water.

\subsubsection{Quadratic Damping}

The effects of the hydrodynamic damping $\mathbf{h}(\bm{\nu})$ over the vehicle, due not only to the
translational but also to rotational motions, can be described in the body-fixed reference frame by:

\begin{equation}
\mathbf{h}(\bm{\nu})=\frac{1}{2}\rho\,[C_{D_x}\upsilon_x|\upsilon_x|,C_{D_y}\upsilon_y|\upsilon_y|,
C_{D_z}\upsilon_z|\upsilon_z|,C_{D_\gamma}\omega_z|\omega_z|]
\label{eq:damp}
\end{equation}

\noindent
where the parameters $C_{D_x}$, $C_{D_y}$, $C_{D_z}$ and $C_{D_\gamma}$ depend on the geometry of the 
vehicle and should be obtained experimentally in a wind tunnel \citep{pinto1}, or on-line estimated 
with adaptive algorithms in a water tank \citep{smallwood3}.

\subsubsection{Added inertia}

Considering that an underwater vehicle typically operates at low speeds, the added inertia matrix, 
$\mathbf{M}_A\in\mathbb{R}^{4\times4}$, could be assumed as diagonally dominant and described as 
follows: 

\begin{equation}
\mathbf{M}_{A}=\mathrm{diag}\,\{C_{M_x}\rho\nabla\:,\:C_{M_y}\rho\nabla\:,\:C_{M_z}\rho\nabla 
\:,\:C_{M_\gamma}\rho\nabla\}
\label{eq:added}
\end{equation}

As with the computation of the hydrodynamic damping, the coefficients $C_{M_x}$, $C_{M_y}$, $C_{M_z}$ 
and $C_{M_\gamma}$ should be determined experimentally. The matrix $\mathbf{M}_A$ must be combined with 
the rigid-body inertia matrix in order to obtain the matrix $\mathbf{M}$ of Eq.~(\ref{eq:movb}).

\section{DYNAMIC POSITIONING SYSTEM}

The dynamic positioning of underwater robotic vehicles is essentially a multivariable control problem.
Nevertheless, as demonstrated by \citet{slotine3}, the variable structure control methodology 
allows different controllers to be separately designed for each degree of freedom (DOF). Over the past 
decades, decentralized control strategies have been successfully applied to the dynamic positioning of 
underwater vehicles \citep{sebastian1,chatcha1,smallwood2,kiriazov1,cunha2,yoerger2}.

Considering that the control law for each degree of freedom can be easily designed with respect to the 
inertial reference frame, Eq.~(\ref{eq:movb}) should be rewritten in this coordinate system.

Remembering that

\begin{equation}
\mathbf{\dot{x}}=\mathbf{J}(\mathbf{x})\bm{\nu}
\label{eq:veli}
\end{equation}

\noindent
where $\mathbf{J}(\mathbf{x})$ is the Jacobian transformation matrix, it can be directly implied that

\begin{equation}
\bm{\nu}=\mathbf{J}^{-1}(\mathbf{x})\mathbf{\dot{x}}
\label{eq:velb}
\end{equation}

\noindent
and

\begin{equation}
\bm{\dot{\nu}}=\mathbf{\dot{J}}^{-1}\mathbf{\dot{x}}+\mathbf{J}^{-1}\mathbf{\ddot{x}}
\label{eq:aceb}
\end{equation}

Therefore, the equations of motion of an underwater vehicle, with respect to the inertial reference
frame, becomes

\begin{equation}
\mathbf{\bar{M}}\mathbf{\ddot{x}}+\mathbf{\bar{k}}+\mathbf{\bar{h}}+\mathbf{\bar{d}}=\bm{\bar{\tau}} 
\label{eq:movi}
\end{equation}

\noindent
where $\mathbf{\bar{M}}=\mathbf{J^\mathrm{-T}M\,J^\mathrm{-1}}$, $\mathbf{\bar{k}}=\mathbf{J^\mathrm{-T}k}
+\mathbf{J^\mathrm{-T}M\,\dot{J}^\mathrm{-1}\dot{x}}$, $\mathbf{\bar{h}}=\mathbf{J^\mathrm{-T}h}$, 
$\mathbf{\bar{d}}=\mathbf{J^\mathrm{-T}d}$ and $\bm{\bar{\tau}}=\mathbf{J}^\mathrm{-T}\bm{\tau}$.

In order to develop the control law with a decentralized approach, Eq.~(\ref{eq:movi}) can be rewritten as 
follows:

\begin{equation}
\ddot{x}_i=\bar{m}_i^{-1}(\bar{\tau}_i-\bar{k}_i-\bar{h}_i-\bar{d}_i);\quad i=1,2,3,4,
\label{eq:mov}
\end{equation}

\noindent
where $x_i$, $\bar{\tau}_i$, $\bar{k}_i$, $\bar{h}_i$ and $\bar{d}_i$ are the components of $\mathbf{x}
=[x,y,z,\gamma]$, $\bm{\bar{\tau}}$, $\mathbf{\bar{k}}$, $\mathbf{\bar{h}}$ and $\mathbf{\bar{d}}$, 
respectively. Concerning $\bar{m}_i$, it represents the main diagonal terms of $\mathbf{J}^\mathrm{-T}
\mathbf{M\,J}^{-1}$. The off-diagonal terms of $\mathbf{J}^\mathrm{-T}\mathbf{M\,J}^{-1}$ are incorporated 
in the vector $\mathbf{\bar{d}}$.

For notational simplicity the index $i$ will be suppressed in Eq.~(\ref{eq:mov}) and, in this way, the equation 
of motion for each degree of freedom (DOF) becomes:

\begin{equation}
\ddot{x}=\bar{m}^{-1}(\bar{\tau}-\bar{k}-\bar{h}-\bar{d})
\label{eq:rov}
\end{equation}

On this basis, according to Eq.~(\ref{eq:law}) and considering $\varepsilon=\dot{\tilde{x}}+\lambda_i\tilde{x}$,
the following could be proposed for each DOF:

\begin{equation}
\bar{\tau}=\bar{k}+\bar{h}+\bar{d}+\bar{m}\left(\ddot{x}_{d}-\lambda\dot{\tilde{x}}\right)-\kappa\varepsilon
\label{eq:lei1}
\end{equation}

But since disturbances are unknown, in this work the value of $d$ will be estimated using a three-layer artificial 
neural network.

\begin{equation}
\hat{d}=\mathbf{W}^\mathrm{T}\sigma(\mathbf{V^\mathrm{T}}\bm{\theta})
\label{eq:ann}
\end{equation}

\noindent
where $\mathbf{W}$ and $\mathbf{V}$ are the weight matrices in, respectively, hidden and output layers. The vector 
$\bm{\theta}$ represents the network input. Standard sigmoid functions are used in hidden layer and the output layer 
has a linear activation function. The weights could be online updated using the conventional backpropagation scheme. 

In this way, the resulting control law could be written as follows:

\begin{equation}
\bar{\tau}=\bar{k}+\bar{h}+\hat{\bar{d}}+\bar{m}\left(\ddot{x}_{d}-\lambda\dot{\tilde{x}}\right)-\kappa\varepsilon
\label{eq:lei2}
\end{equation}

\section{SIMULATION RESULTS}

\begin{figure}[htb]
\centering
\mbox{
\subfigure[Tracking error $e$.]{\label{fg:erro1} 
\includegraphics[width=0.47\textwidth]{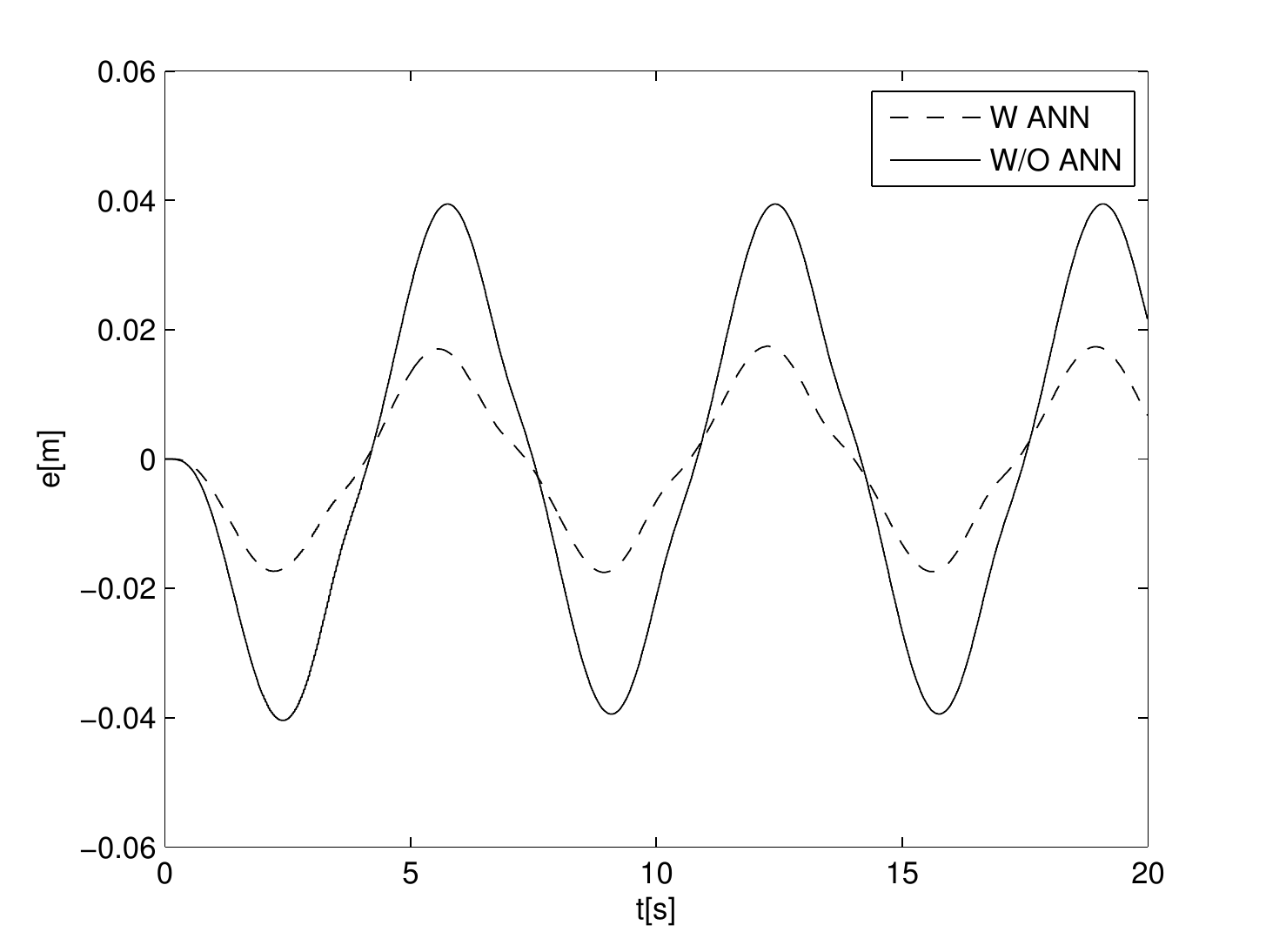}}
\subfigure[Thrust force $u$.]{\label{fg:cont1} 
\includegraphics[width=0.47\textwidth]{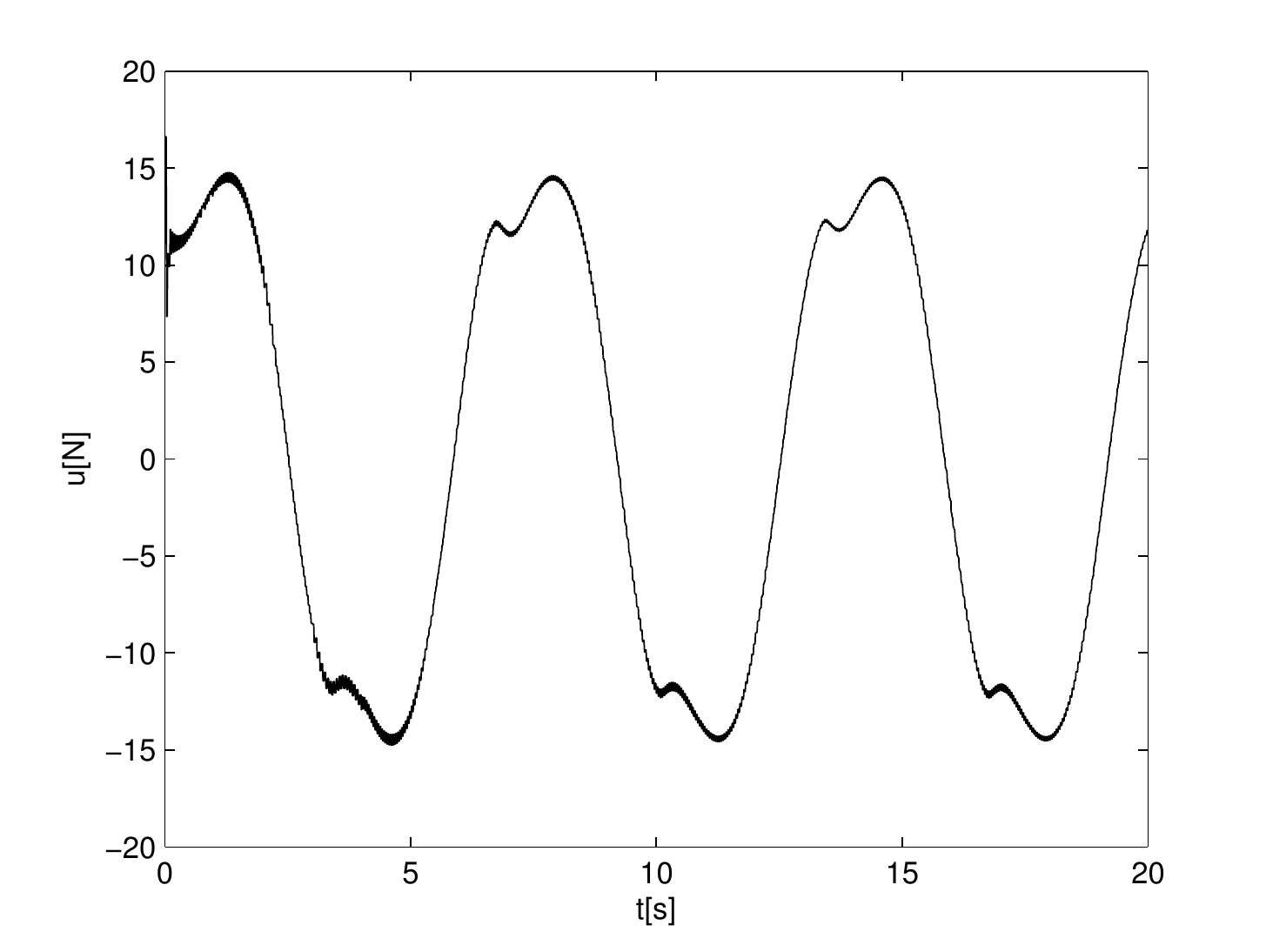}}
}
\caption{Tracking with $z_d=0.5[1-\cos(0.1\pi t)]$ and $\mathbf{\tilde{z}}(0)=\mathbf{0}$.}  
\label{fg:sim1}
\end{figure}

\begin{figure}[htb]
\centering
\mbox{
\subfigure[Trajectory Tracking.]{\label{fg:traj2} 
\includegraphics[width=0.47\textwidth]{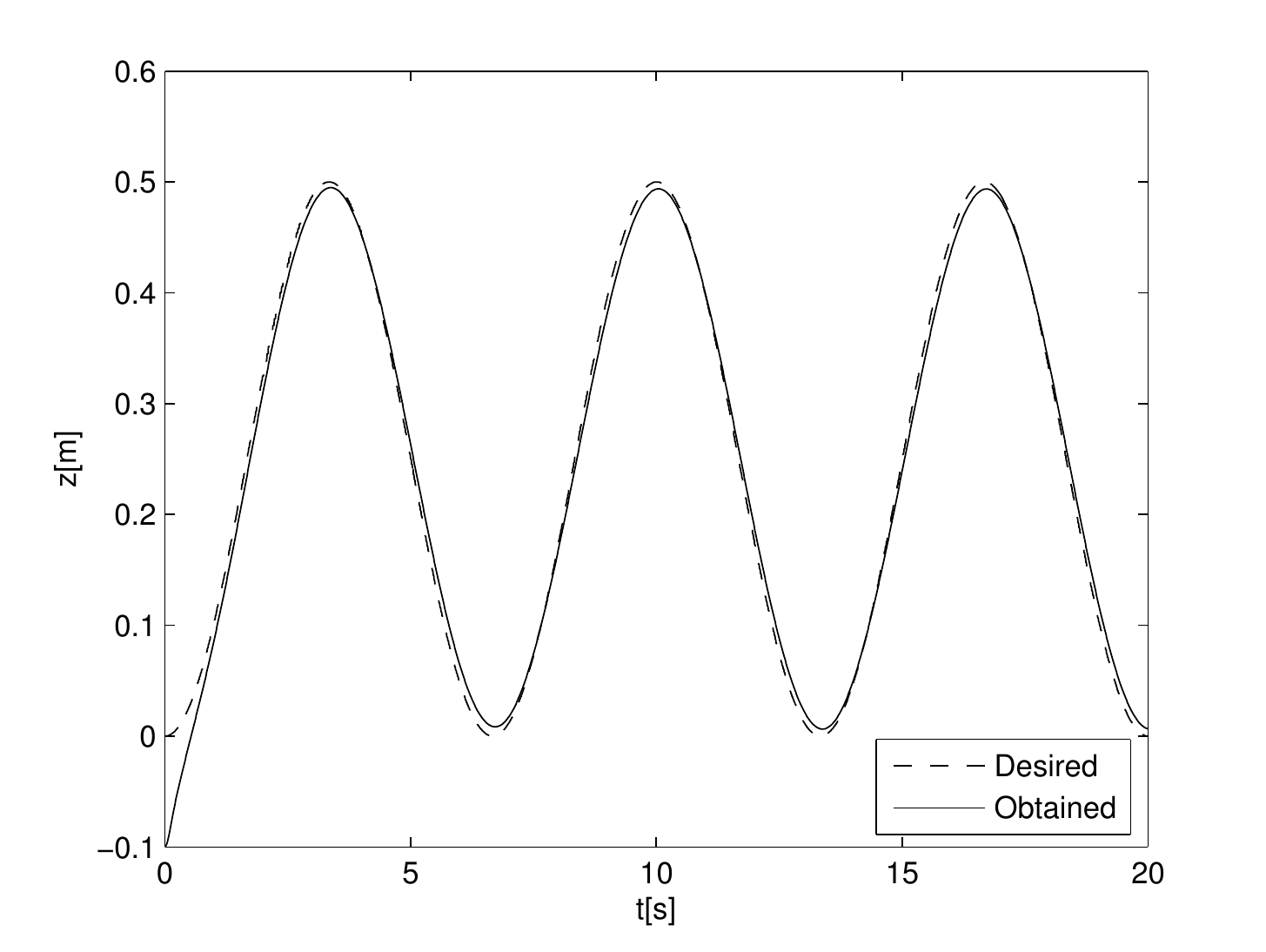}}
\subfigure[Phase portrait of the tracking error.]{\label{fg:erro2} 
\includegraphics[width=0.47\textwidth]{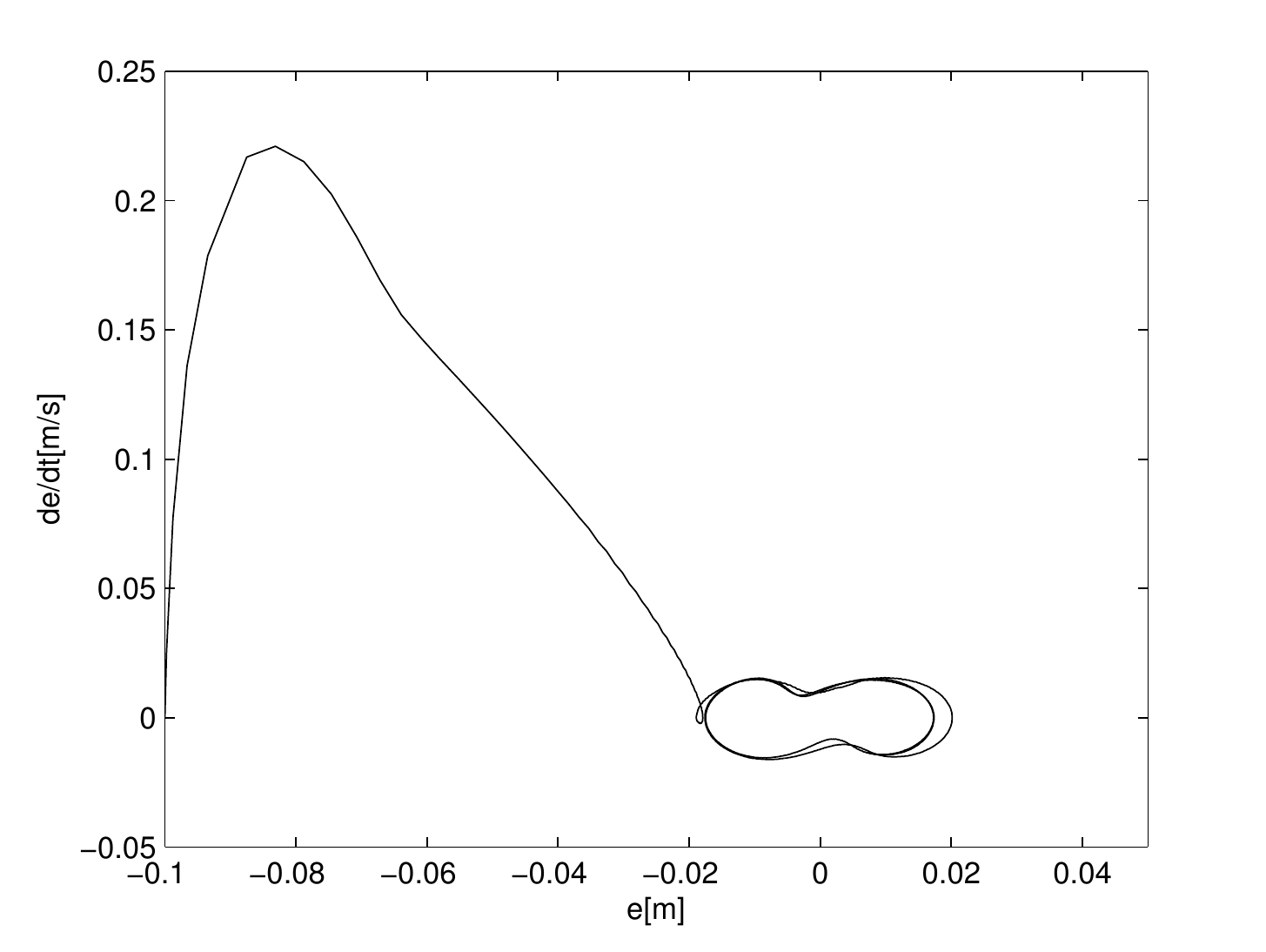}}
}
\caption{Tracking with $\mathbf{\tilde{z}}(0)=[0.1,0.0]$ and ANN compensation.}  
\label{fg:sim2}
\end{figure}

\begin{figure}[htb]
\centering
\mbox{
\subfigure[Trajectory Tracking.]{\label{fg:traj3} 
\includegraphics[width=0.47\textwidth]{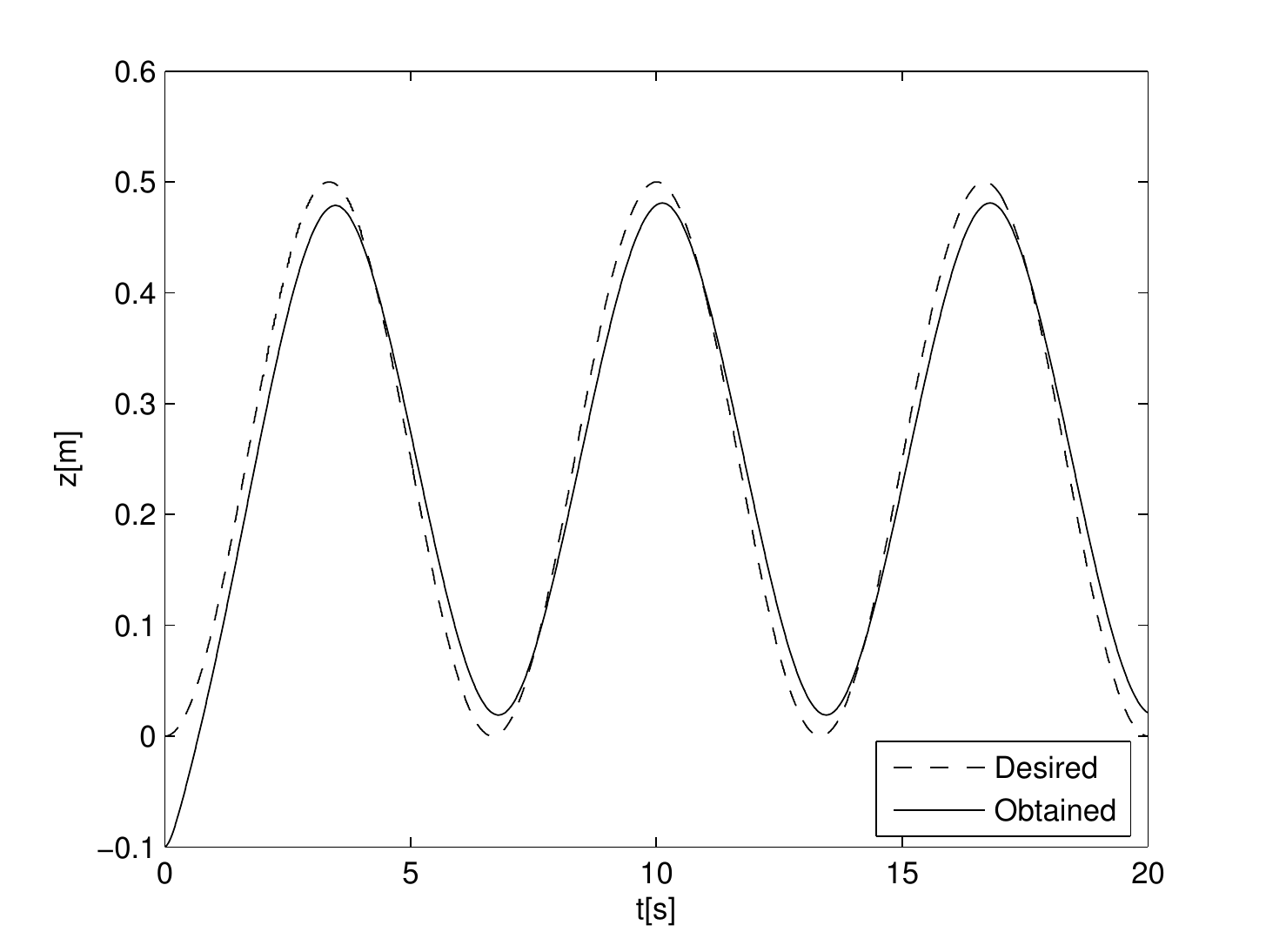}}
\subfigure[Phase portrait of the tracking error.]{\label{fg:erro3} 
\includegraphics[width=0.47\textwidth]{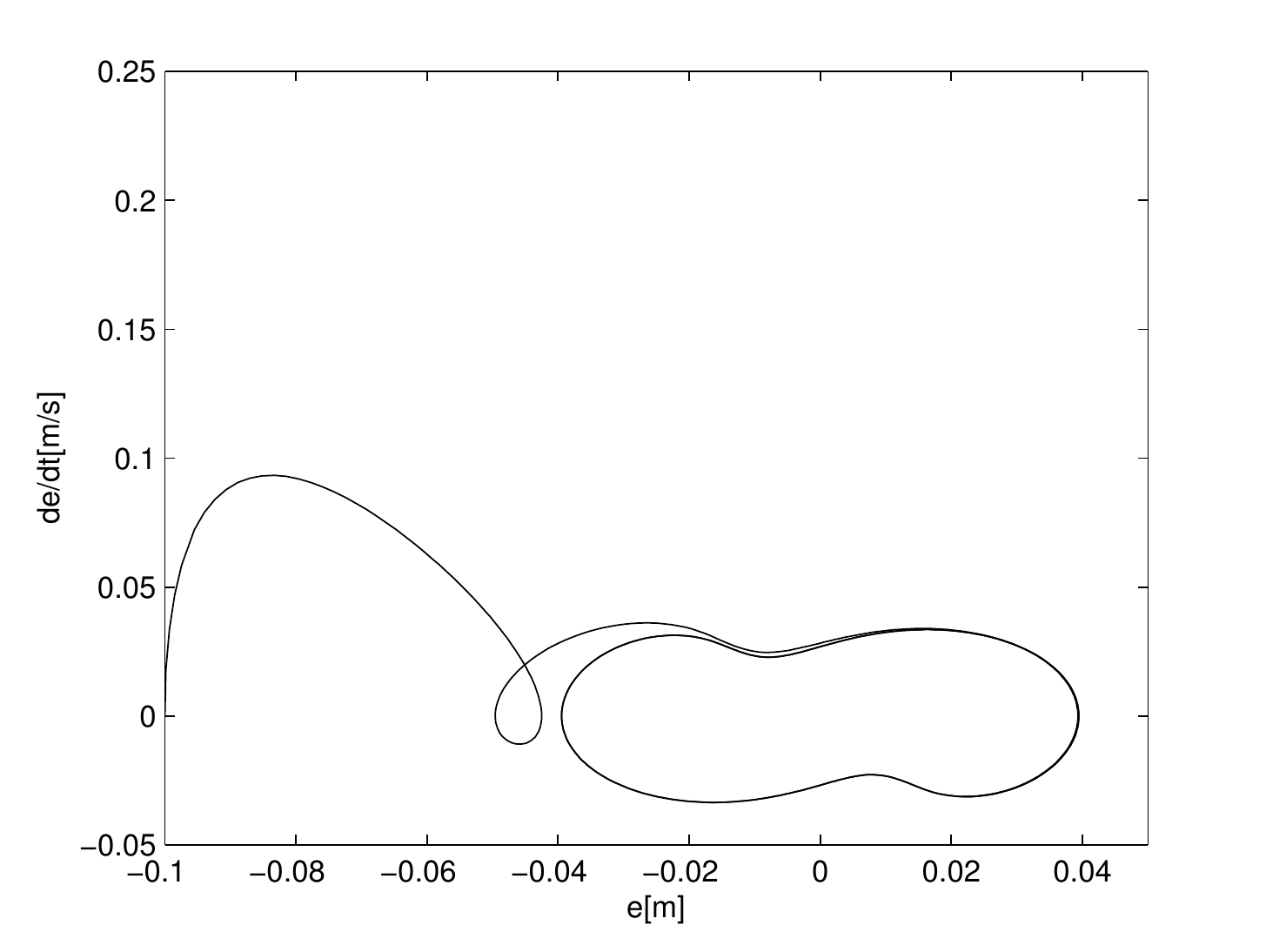}}
}
\caption{Tracking with $\mathbf{\tilde{z}}(0)=[0.1,0.0]$ but without ANN compensation.}  
\label{fg:sim3}
\end{figure}

\begin{figure}[htb]
\centering
\mbox{
\subfigure[Trajectory Tracking.]{\label{fg:traj4} 
\includegraphics[width=0.47\textwidth]{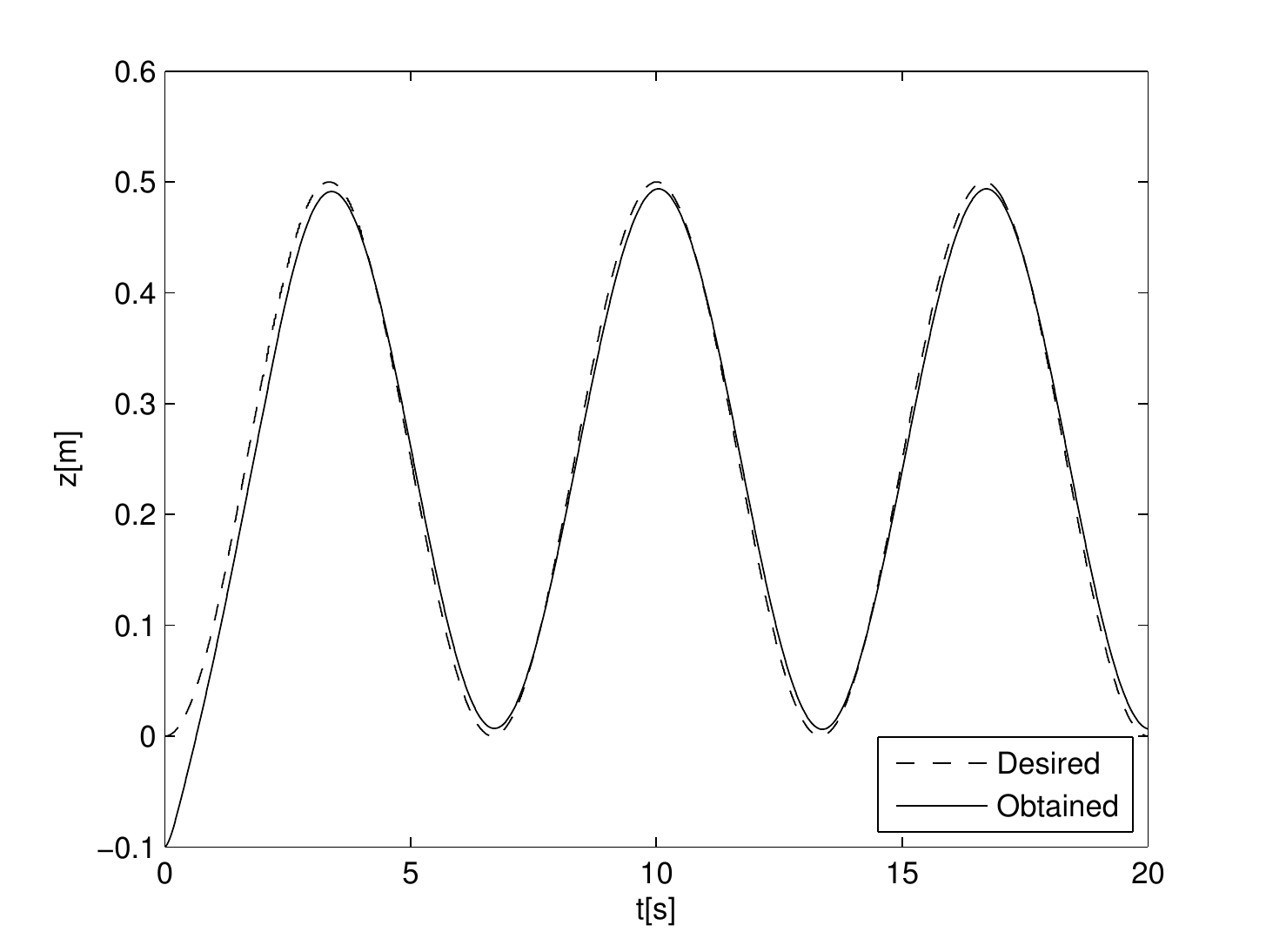}}
\subfigure[Phase portrait of the tracking error.]{\label{fg:erro4} 
\includegraphics[width=0.47\textwidth]{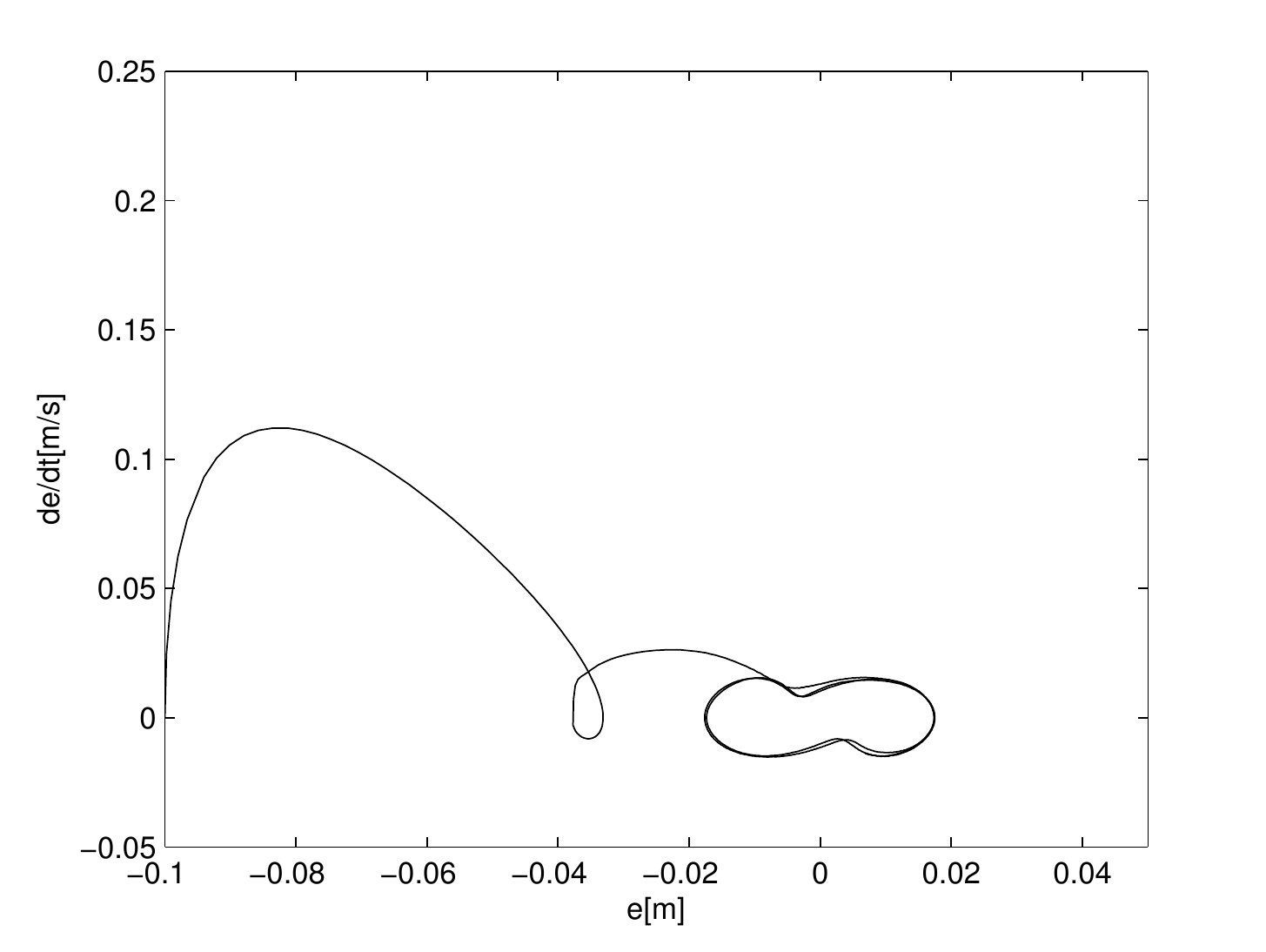}}
}
\caption{Tracking with $\mathbf{\tilde{z}}(0)=[0.1,0.0]$ and ANN compensation starting after 2 s.}  
\label{fg:sim4}
\end{figure}

In order to evaluate the control system performance, three different numerical simulations were performed for the depth
regulation of an underwater robotic vehicle. The obtained results were presented from Fig.~\ref{fg:sim1} to Fig.~\ref{fg:sim4}.

In the first case, it was considered that the initial state coincides with the initial desired state, $\mathbf{\tilde{z}}(0)=
[\tz(0),\dtz(0)]^\mathrm{T}=\mathbf{0}$. Figure~\ref{fg:sim1} gives the corresponding results for the tracking of $z_d=0.5[1-
\cos(0.1\pi t)]$. Regarding controller and model parameters, the following values were chosen $\bar{m}=50$~kg, $\bar{h}=250$ 
Ns$^2$/m$^2$, $\kappa=3.5$ and $\lambda=3.5$. The capability of the proposed scheme to deal with uncertainties was appraised 
by choosing the parameters for the controller based on the assumption that exact values are not known but with a maximal 
uncertainty of $\pm10\%$ over adopted values for the model parameters. 

As observed in Fig.~\ref{fg:sim1}, the proposed control scheme allows the underwater robotic vehicle to track the desired 
trajectory with a small tracking error. Through the comparative analysis showed in Fig.~\ref{fg:erro1}, the improved 
performance of the proposed controller over the uncompensated counterpart can be easily ascertained. 

In the second simulation the initial state and initial desired state are not equal, $\mathbf{\tilde{z}}(0)=[0.1,0.0]$. The 
controller and model parameters and the desired trajectory were defined as before. Figures~\ref{fg:sim2}~and~\ref{fg:sim3}
show the corresponding results, respectively, with and without the neural network compensation scheme.

As observed in Fig.~\ref{fg:traj2}, even if the initial state and initial desired state are not equal, the proposed control 
scheme allows the trajectory tracking with a small tracking error. By comparing Fig.~\ref{fg:erro2} and Fig.~\ref{fg:erro3}, 
it can be verified that the adoption of a neural-network compensation scheme leads to a smaller limit cycle in the phase
portrait related to the tracking error. Although the compensation scheme improves the tracking performance, a larger overshoot
with respect to velocity error can be seen in Fig.~\ref{fg:erro2}. This overshoot is due to the fact that the initial tracking
error is different of zero. In order to avoid this phenomenon, we will let the neural-network compensator to wait for 2 seconds
until it starts. Figure~\ref{fg:sim4} shows the obtained results.

\section{CONCLUDING REMARKS}

In this paper, a nonlinear controller was proposed to deal with the dynamic positioning system for underwater robotic vehicles. 
To enhance the tracking performance the adopted strategy embedded an artificial neural-network based scheme within the conventional
nonlinear controller for uncertainty/disturbance compensation. The stability and convergence properties of the closed-loop systems 
were analytically proven using Lyapunov stability theory. Through numerical simulations, the improved performance over the uncompensated 
control strategy was demonstrated.

\section{ACKNOWLEDGEMENTS}

The authors would like to acknowledge the support of the Brazilian National Research Council (CNPq), the Brazilian Coordination for the 
Improvement of Higher Education Personnel (CAPES) and the German Academic Exchange Service (DAAD).

\end{document}